\documentclass[a4paper]{svproc}
\usepackage{amssymb,amsmath}
\usepackage{prettyref}
\usepackage{mathrsfs}
\usepackage{graphicx}
\usepackage{wrapfig}
\usepackage{xcolor}
\usepackage{subfig}
\usepackage{multirow}
\usepackage{booktabs}
\usepackage{algorithm}
\usepackage{algpseudocode}
\usepackage[utf8]{inputenc}
\usepackage[english]{babel}
\usepackage[nomessages]{fp}

\newrefformat{Fig}{Figure~\ref{#1}}
\newrefformat{fig}{Figure~\ref{#1}}
\newrefformat{par}{Section~\ref{#1}}
\newrefformat{appen}{Appendix~\ref{#1}}
\newrefformat{sec}{Section~\ref{#1}}
\newrefformat{sub}{Section~\ref{#1}}
\newrefformat{table}{Table~\ref{#1}}
\newrefformat{alg}{Algorithm~\ref{#1}}
\newrefformat{Alg}{Algorithm~\ref{#1}}
\newrefformat{Def}{Definition~\ref{#1}}
\newrefformat{Thm}{Theorem~\ref{#1}}
\newrefformat{Lem}{Lemma~\ref{#1}}
\newrefformat{step}{Step~\ref{#1}}
\newrefformat{ln}{Line~\ref{#1}}
\newrefformat{eq}{Equation~\ref{#1}}
\newrefformat{pb}{Problem~\ref{#1}}
\newrefformat{it}{Item~\ref{#1}}
\newrefformat{te}{Term~\ref{#1}}
\def\Eqref Eq:#1:{\eqref{eq:#1}}
\newrefformat{Eq}{Equation~\Eqref#1:}

\newcommand{\E}[1]{\mathbf{#1}}
\newcommand{\TE}[1]{\textbf{#1}}

\newcommand{\TWO}[2]{\left(\setlength{\arraycolsep}{1pt}\begin{array}{cc}{#1}, & {#2}\end{array}\right)}
\newcommand{\TWOC}[2]{\left(\setlength{\arraycolsep}{1pt}\begin{array}{c}#1 \\ #2\end{array}\right)}

\newcommand{\argcon}{\E{s.t.}\quad\;\;}
\newcommand{\argmin}[1]{\underset{#1}{\E{argmin}}\quad}
\newcommand{\argminP}[1]{\E{argmin}\quad}

\newcommand{\argmaxP}[1]{\E{argmax}\quad}

\newcommand{\NN}{\E{n}}

\newcommand{\BO}{\mathcal{O}}
\newcommand{\SOSA}{\mathcal{SOS}_1}
\newcommand{\SOSB}{\mathcal{SOS}_2}
\newcommand{\CNN}[3]{\NN_{{#1\rightarrow #2#3}}}
\definecolor{motor}{HTML}{289942}
\definecolor{fixed}{HTML}{D02824}
\definecolor{movable}{HTML}{000000}
\definecolor{endeffector}{HTML}{3AB4E5}
\newcommand{\GAMMA}[5]{
\FPeval{\X}{cos((#1*45+45/2)*pi/180)*#4-#2}
\FPeval{\Y}{sin((#1*45+45/2)*pi/180)*#4+#3}
\put(\X,\Y){#5}
}
\newcommand{\GAMMAI}[5]{
\FPeval{\X}{cos((#1*45)*pi/180)*#4/2-#2}
\FPeval{\Y}{sin((#1*45)*pi/180)*#4/2+#3}
\put(\X,\Y){#5}
}
\newcommand{\TR}[1]{\textcolor{fixed}{#1}}

\usepackage{url}

\usepackage{amsthm}

\begin{document}
\mainmatter              
\title{Globally Optimal Joint Search of Topology and Trajectory for Planar Linkages}
\titlerunning{Joint Search of Topology and Trajectory for Planar Linkages}
\author{Zherong Pan\inst{1} \and Min Liu\inst{2,4} \and Xifeng Gao\inst{3} \and Kai Xu\inst{2} \and Dinesh Manocha\inst{4}}
\authorrunning{Pan et al.} 
%
\tocauthor{Zherong Pan, Min Liu, Xifeng Gao, Dinesh Manocha}
\institute{
Department of Computer Science, 
University of North Carolina, North Carolina NC 27514, USA,\\
\email{zherong@cs.unc.edu}\\
\and
School of Computer, National University
of Defense Technology, Hunan HN 410073, China,\\
\email{gfsliumin@gmail.com, kevin.kai.xu@gmail.com}\\
\and
Department of Computer Science,
Florida State University, Florida FL 32306, USA,\\
\email{gao@cs.fsu.edu}\\
\and
Department of Computer Science and Electrical \& Computer Engineering, 
University of Maryland at College Park, Maryland MD 20742, USA,\\
\email{dm@cs.umd.edu}
}

\maketitle              

\begin{abstract}
We present a method to find globally optimal topology and trajectory jointly for planar linkages. Planar linkage structures can generate complex end-effector trajectories using only a single rotational actuator, which is very useful in building low-cost robots. We address the problem of searching for the optimal topology and geometry of these structures. However, since topology changes are non-smooth and non-differentiable, conventional gradient-based searches cannot be used. We formulate this problem as a mixed-integer convex programming (MICP) problem, for which a global optimum can be found using the branch-and-bound (BB) algorithm. Compared to existing methods, our experiments show that the proposed approach finds complex linkage structures more efficiently and generates end-effector trajectories more accurately.
\vspace{-5px}
\end{abstract}
\section{Introduction}
\begin{wrapfigure}{r}{6cm}
\centering
\vspace{-15px}
\includegraphics[width=0.4\textwidth]{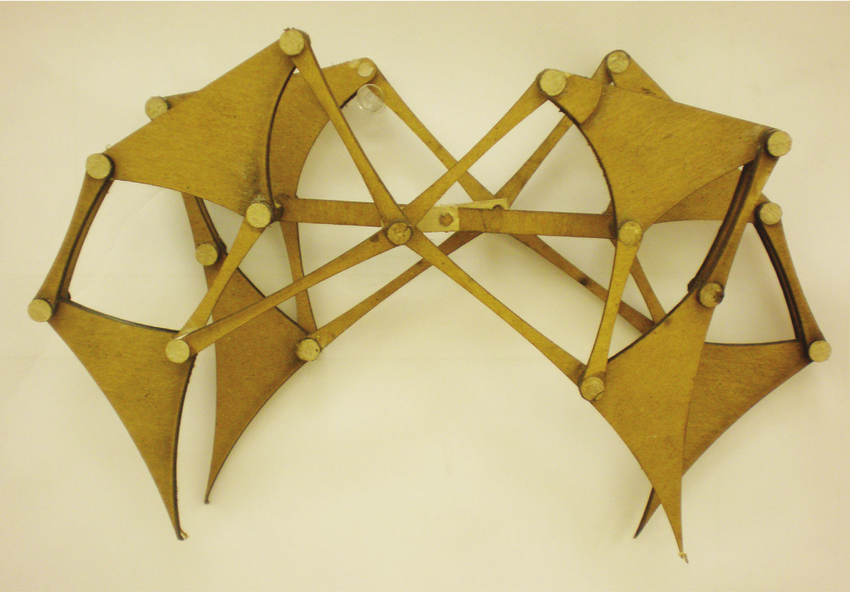}
\vspace{-5px}
\caption{\label{fig:strandbeest} An example of planar linkages, used in a strandbeest robot for 2D walking. See \cite{nansai2013dynamic} for more details.}
\vspace{-10px}
\end{wrapfigure}

A planar linkage is a mechanical structure built with a set of rigid bodies connected by hinge joints. This structure typically has one effective degree-of-freedom actuated by a rotational motor. Since they impose a minimal burden on controller design, these structures are widely used as building blocks for low-cost toys and robots, as illustrated in \prettyref{fig:strandbeest}. By combining a series of hinge joints, the end-effector of the planar linkage will trace out a complex curve that can fulfill various requirements of different types of locomotion, including walking and swimming \cite{HERNANDEZ2016AHO,Thomaszewski:2014:CDL:2601097.2601143}.

A challenging problem in mechanics design is to find the linkage structure with an end-effector that will trace out a given curve. This problem is challenging in that it searches over three coupled variables: topology, geometry, and trajectory. The linkage topology determines which rigid bodies are connected and the order of their connections. Clearly, the topology is a non-smooth and non-differentiable decision variable. The linkage geometry determines the shape of each rigid link. Finally, the trajectory determines the pose of the linkage structure at each time instance. The last two variables are smooth and differentiable, but directly optimizing them induces non-convex functions. Previous works \cite{ha2018computational,Zhu:2012:MMT:2366145.2366146} have proposed various solutions to address problems of this kind. These methods rely on random searches, such as $A^*$ \cite{ha2018computational} and covariance matrix adaptation \cite{Zhu:2012:MMT:2366145.2366146}, to try different topologies. Then, for each topology, they perform non-linear programming (NLP) under the given topology to determine the geometry and trajectory. However, these methods are computationally expensive because a huge number of samples are needed for the random search to converge. Moreover, even after determining the topology, these methods can find only sub-optimal solutions due to the non-convex nature of NLP.

\setlength{\columnsep}{10pt}
\TE{Main Results:} Given the input of a target trajectory, we present a new method that can efficiently compute a planar linkage structure with globally optimal topology and geometry configurations and an accurate trajectory reproduction. Based on recent advances in mixed-integer modeling \cite{vielma2015mixed,dai2017global,trespalacios2015improved}, we relax this joint search problem as an MICP problem, the global optimum of which is arbitrarily close to the global optimum of the original problem. The main benefit of MICP relaxation is that the search can be accomplished efficiently using the BB algorithm. BB is more strategic than random search, as used by \cite{Zhu:2012:MMT:2366145.2366146}, because it cuts impossible or sub-optimal search spaces at an early stage, leading to higher efficiency. We have compared MICP with prior methods using different examples. The results show that our proposed MICP approach finds solutions more efficiently and that the resulting structure matches the target trajectory more closely.

In the rest of the paper, we first review related work in \prettyref{sec:related} and then formulate our joint search problem in \prettyref{sec:problem}. The MICP model and various constraints required for the integrity of the planar linkage are presented in \prettyref{sec:method}. Results and the evaluation of the proposed approach are given in \prettyref{sec:results}.
\section{\label{sec:related}Related Work}
In this section, we review related work in robot design optimization, mixed-integer modeling, and topology optimization.

\TE{Robot Design Optimization:} Robot design optimization is a superset of conventional topology and truss optimization \cite{LIU2016161} where the decision variables are only topology or geometry. This is because the specification of a robot design is given as a movement pattern \cite{Ha2017JointOO}, leading to a joint search in the space-time domain. The joint search problem greatly expands the search space. As a result, many prior methods do not work since they only optimize a subset of decision variables \cite{Ha2017JointOO,Thomaszewski:2014:CDL:2601097.2601143,bacher2015linkedit,saar2018model,spielberg2017functional}. Recent works \cite{Zhu:2012:MMT:2366145.2366146,ha2018computational,song2017computational} search for all variables simultaneously. However, these methods are based on random search techniques, which usually require a large amount of trial and error and find sub-optimal solutions. 

\TE{Mixed-Integer Modeling:} The main benefit of mixed-integer modeling is the use of the well-studied BB algorithm \cite{lawler1966branch}. BB allows us to find the global optimum of non-convex programming problems, while only visiting a small fraction of the search space. Mixed-integer models have been applied to a large variety of problems including motion planning \cite{ding2011mixed}, inverse kinematics \cite{dai2017global}, network flows \cite{conforti2009network}, and mesh generations \cite{bommes2009mixed}. By applying the big-M method \cite{trespalacios2015improved}, McCormick envelopes and piecewise approximations \cite{liberti2004reformulation}, and general non-convex problems can be easily relaxed as MICP problems. Prior works \cite{kanno2013topology,lobato2003mixed} have also formulated topology optimization problems as MICP. However, our work is the first to formulate the planar linkage problem as MICP and we employ MICP to concurrently find the optimal topology, geometry, and trajectory of a linkage.

\TE{Topology Optimization:} Topology optimization of a continuum is a well-studied problem \cite{LIU2016161}. An efficient algorithm can smoothen the problem and use gradient-based method to search for locally optimal structures over a search space of millions of dimensions. This technique has been widely used in the design of soft robots \cite{zhang2017design,zhang2018design,zhu2017two}. However, the optimization of articulated robots is more challenging because the optimized structure must satisfy the joint constraints, making the decision variable non-smooth. Existing techniques use mixed-integer \cite{kanno2013topology,lobato2003mixed} or random search techniques \cite{Zhu:2012:MMT:2366145.2366146,song2017computational} to optimize over these decision variables.
\section{\label{sec:problem}Joint Search for Planar Linkages}
\begin{figure}[ht]
\centering
\vspace{-5px}
\begin{tabular}{@{}c@{}}
\includegraphics[width=0.25\textwidth]{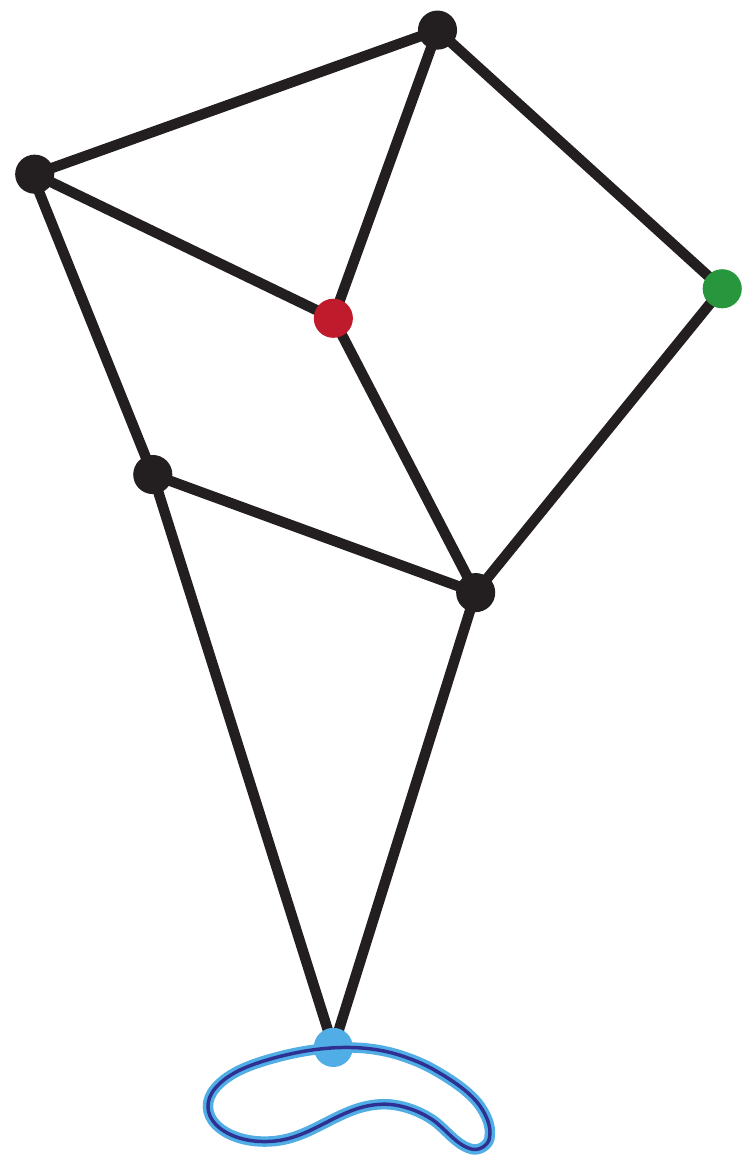}
\put(  2,100){\textcolor{motor}{$\NN_1$}}
\put(-43,95 ){\textcolor{fixed}{$\NN_2$}}
\put(-37,137){\textcolor{movable}{$\NN_3$}}
\put(-97,111){\textcolor{movable}{$\NN_4$}}
\put(-27,63 ){\textcolor{movable}{$\NN_5$}}
\put(-83,78 ){\textcolor{movable}{$\NN_6$}}
\put(-42,16 ){\textcolor{endeffector}{$\NN_7$}}
\end{tabular}
\begin{tabular}{@{}c@{}}
\includegraphics[width=0.4\textwidth]{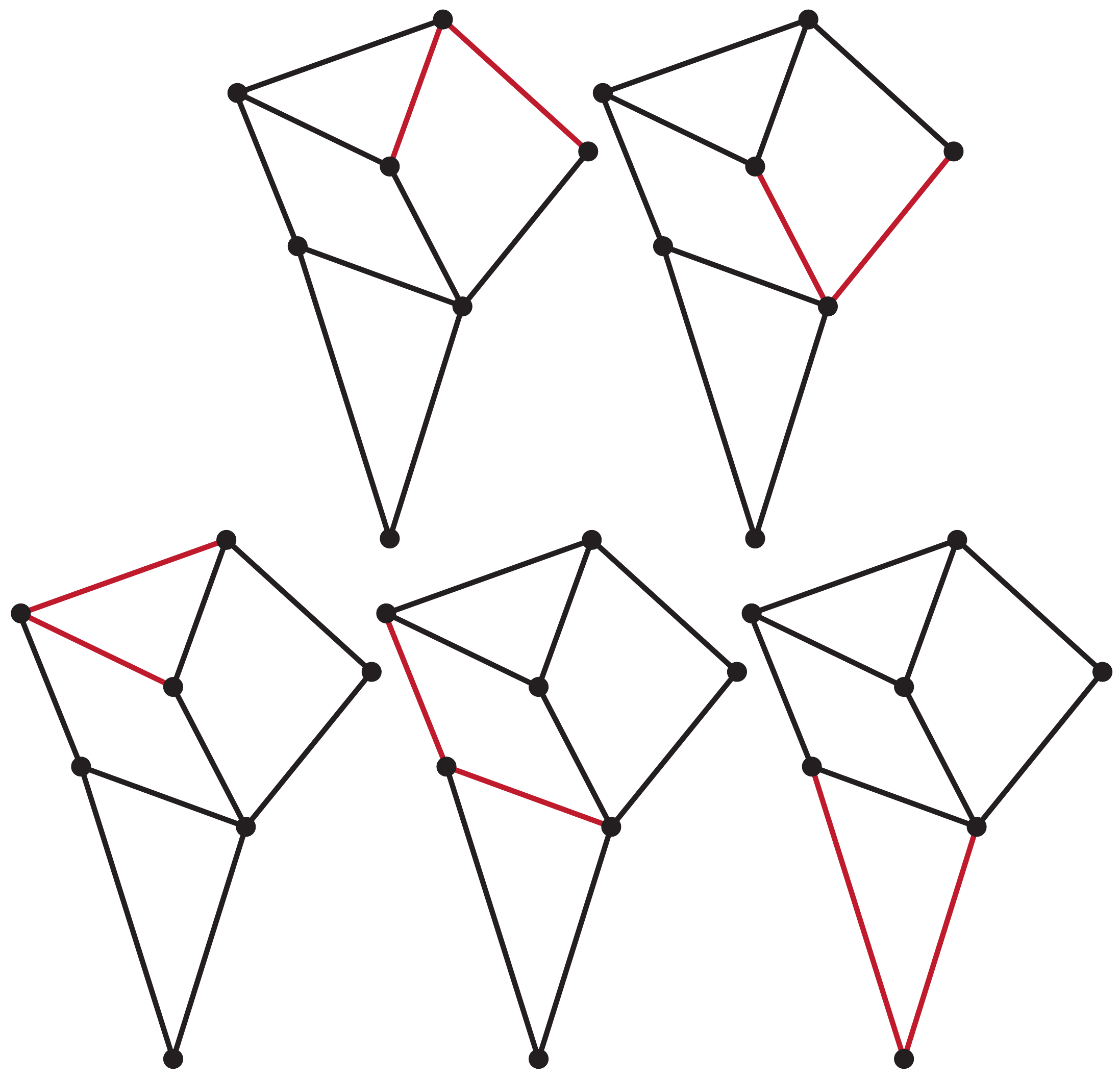}
\put(-80 ,75){$\CNN{3}{2}{1}$}
\put(-35 ,75){$\CNN{5}{2}{1}$}
\put(-107,10){$\CNN{4}{3}{2}$}
\put(-62,10){$\CNN{6}{5}{4}$}
\put(-17,10){$\CNN{7}{6}{5}$}
\end{tabular}
\caption{\label{fig:linkage} (a): The Jansen's mechanics used in \prettyref{fig:strandbeest} is a planar linkage structure involving 7 nodes. The motor node $\NN_1$ is green, the fixed node $\NN_2$ is red, the movable nodes $\NN_{3,4,5,6}$ are black, and the end-effector node $\NN_7$ is blue. Our goal is to find the topology and geometry of the linkage such that the end-effector curve matches the blue target curve. (b): Our MICP formulation is based on the prior symbolic representation \cite{kecskemethy1997symbolic,bacher2015linkedit}. This representation assumes that each node is connected to exactly two other nodes with lower indices: $\CNN{3}{2}{1}$, $\CNN{5}{2}{1}$, $\CNN{4}{3}{2}$, $\CNN{6}{5}{4}$, $\CNN{7}{6}{5}$.}
\vspace{-10px}
\end{figure}
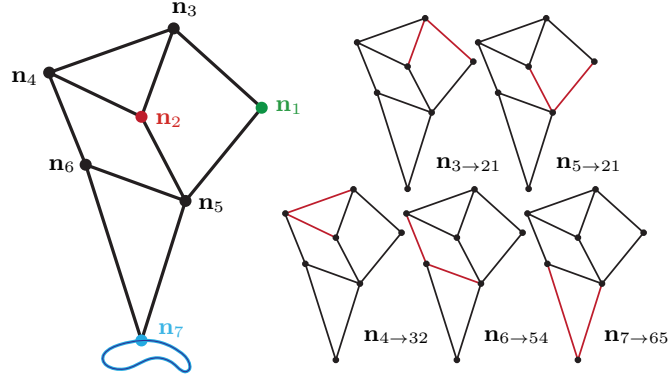
In this section, we introduce the problem of joint searches for planar linkages. Our problem is to search for a structure, as illustrated in \prettyref{fig:linkage}a, where we have a set of rod-like rigid bodies connected with each other using hinge joints. As a result, the end points of these rigid bodies can take at most $N$ distinct positions, denoted as node set: $\NN_{1,\cdots,N}$. Of these nodes, $\NN_1$ is the rotational motor and $\NN_N$ is the end-effector. Within one limit cycle, $\NN_1$ follows a circular curve centered at $\TWO{X_C}{Y_C}$ with a radius $R$:
\begin{align}
\label{eq:motor}
\NN_1(t)=\TWO{\E{sin}(t)R+X_C}{\E{cos}(t)R+Y_C},
\end{align}
which induces trajectories of other nodes $\NN_i(t)$ via forward kinematics. The other $N-2$ nodes can be one of two kinds: fixed or movable. In addition, a rigid body may exist between each pair of nodes $\NN_{i,j}$, in which case $\|\NN_i(t)-\NN_j(t)\|$ is a constant.

Given these definitions, the input to our problem is a target end-effector trajectory $\NN_N^*(t)$. The output of our method is the following set of variables defining both the topology and geometry of a planar linkage:
\begin{itemize}
    \item An integer vector of size $N$ (the number of nodes), which containing the type of each node: fixed or movable.
    \item An $N\times N$ symmetric binary matrix $C^{N\times N}$ where $C_{ij}=1$ means a rigid body connects $\NN_{i,j}$.
    \item The position of $\NN_{1,\cdots,N}(t)$ at a certain, arbitrary time instance $t$.
\end{itemize}
The goal of our method is to find the globally optimal set of variables that minimizes $\int \|\NN_N(t)-\NN_N^*(t)\|^2 dt$.
\section{\label{sec:method}MICP Formulation of Joint Search}
In this section, we present a set of linear constraints and quadratic objective functions for relaxing the joint search as an MICP problem. We first introduce the set of topology constraints to ensure the well-posed nature of the structure in \prettyref{sec:topology} and then present constraints and objective functions for geometric correctness in \prettyref{sec:geometry}.

\subsection{\label{sec:topology}Topology Constraints}
As illustrated in \prettyref{fig:linkage}b, our method is based on the symbolic representation presented in \cite{kecskemethy1997symbolic,Thomaszewski:2014:CDL:2601097.2601143}, which assumes that each movable node is attached to two other nodes. These nodes can be of any type but must have lower node indices. As a result, forward kinematics can be processed sequentially even on linkage structures with closed loops. 

Since the number of nodes is unknown, we assume that the maximal number of nodes is $K>N$. For each node other than the first motor node $\NN_1$, we need a binary variable $U_i$ such that $U_i=1$ indicates $\NN_i$ is used as a part of the planar linkage structure. In addition, we need another binary variable $F_i$ such that $F_i=1$ indicates $\NN_i$ is fixed and $F_i=0$ indicates $\NN_i$ is movable. These two sets of variables are under the constraint that only a used node can be movable. In addition, we assume that the last node $\NN_K$ is the end-effector that must be used. In summary, we introduce the following sets of variables and node-state constraints:
\begin{equation}
\begin{aligned}
\label{eq:state}
\exists U_i,F_i\in\{0,1\} &\quad\forall i=1,\cdots,K \\
1-F_i\leq U_i&    \\
U_1=U_K=1&  \\
F_1=0&.
\end{aligned}
\end{equation}

Our next set of constraints ensures local topology correctness. It ensures that each movable node is connected to exactly two other nodes with lower indices. As a result, the movable node and the two other nodes will form a triangle and the position of the movable node can then be determined via the Law of Cosine \cite{Ha2017JointOO}. We introduce auxiliary variables $C_{ji}^1$ to indicate whether $\NN_j$ is the first node to which $\NN_i$ is connected. $C_{ji}^2$ indicates whether $\NN_j$ is the second node to which $\NN_i$ is connected. In addition, we introduce two verbose variables $C_{0i}^{1,2}=1$ to indicate that $\NN_i$ is connected to nothing. The resulting constraint set is:
\begin{equation*}
\begin{aligned}
\exists C_{ji},C_{ji}^1,C_{ji}^2\in\{0,1\}&\quad\forall j,i=1,\cdots,K\land j<i   \\
C_{ji}=C_{ji}^1+C_{ji}^2&   \\
C_{ji}^1\leq U_j\land C_{ji}^2\leq U_j&   \\
\sum_{j=1}^{i-1} C_{ji}=2-2F_i&\quad\forall i=2,\cdots,K \\
\exists C_{0i}^d\in\{0,1\}&\quad \forall d=1,2  \\
\sum_{j=0}^{i-1} C_{ji}^d=1&.
\end{aligned}
\end{equation*}
When $\NN_i$ is fixed in the above formulation, then $F_i=1$ in \prettyref{eq:connectivity} and all $C_{ji}$ are zero except for $C_{0i}^{1,2}=1$ due to the sum-to-one constraints. If $\NN_i$ is movable, then $F_i=0$ and $C_{ji}$ sums to two. As a result, there must be $j_1,j_2<i$ such that $C_{j_1i}^1=1$ and $C_{j_2i}^2=1$. Note that $j_1$ and $j_2$ must be different because otherwise the constraint that $C_{ji}\in[0,1]$ will be violated. In addition, since the first node $\NN_1$ is the motor node, it is excluded from these connectivity constraints. However, this naive formulation will require binary variables for each pair of $\NN_j$ and $\NN_i$, which requires $\BO(K^2)$ binary variables all together. Instead, we adopt the idea of special ordered set of type 1 ($\SOSA$) \cite{vielma2011modeling} and model these constraints using $\BO(K\lceil\E{log}K\rceil)$ binary variables. Intuitively, $\SOSA$ constrains that only one variable in a set can take a non-zero value and it can be achieved by using a logarithm number of binary variables. The improved constraint set is:
\begin{equation}
\begin{aligned}
\label{eq:connectivity}
\exists C_{ji},C_{ji}^1,C_{ji}^2\in\TR{[0,1]}&\quad\forall j,i=1,\cdots,K\land j<i   \\
C_{ji}=C_{ji}^1+C_{ji}^2&   \\
C_{ji}^1\leq U_j\land C_{ji}^2\leq U_j&   \\
\sum_{j=1}^{i-1} C_{ji}=2-2F_i&\quad\forall i=2,\cdots,K \\
\exists C_{0i}^d\in\TR{[0,1]}&\quad \forall d=1,2  \\
\TR{\{C_{ji}^d|j=0,\cdots,i-1\}\in\SOSA}&    \\
\sum_{j=0}^{i-1} C_{ji}^d=1&.
\end{aligned}
\end{equation}

Finally, we introduce a third set of constraints to ensure global topology correctness. This set of constraints ensures that the linkage structure contains no wasted structures. In other words, each node must have some influence on the trajectory of the end-effector node and the first motor node must be connected to others. We model these constraints using the MICP formulation of network flows \cite{conforti2009network}. Specifically, each node $\NN_i$ will generate an outward flux that equals to $U_i$, and we assume that there is a flow edge defined between each pair of nodes with capacity $Q_{ji}$. We require inward-outward flux balance for each node except for the end-effector node:
\begin{equation}
\begin{aligned}
\label{eq:balance}
\exists Q_{ji}\in[0,\infty]&\quad\forall j,i=1,\cdots,K\land j<i    \\
Q_{ji}\leq C_{ji}K& \\
U_i+\sum_{j=1}^{i-1} Q_{ji}=\sum_{k=i+1}^{K} Q_{ik} &\quad\forall i=1,\cdots,K-1,
\end{aligned}
\end{equation}
where we adopt the big-M method \cite{trespalacios2015improved} in the second constraint to ensure that only edges between connected nodes can have a capacity up to $K$. Using a similar idea, we also formulate a constraint that a movable node must be connected to at least one other movable node. We assume that each node $\NN_i$ generates a reversed outward flux that equals to $1-F_i$, and we assume that there is a flow edge defined between each pair of nodes with capacity $R_{ji}$. We require inward-outward flux balance for each node except for the motor node:
\begin{equation}
\begin{aligned}
\label{eq:balance2}
\exists R_{ji}\in[0,\infty]&\quad\forall j,i=1,\cdots,K\land j<i    \\
R_{ji}\leq C_{ji}K\land R_{ji}\leq (1-F_j)K& \\
\sum_{j=1}^{i-1} R_{ji}=1-F_i+\sum_{k=i+1}^{K} R_{ik} &\quad\forall i=2,\cdots,K,
\end{aligned}
\end{equation}
These three constraints ensure that the planar linkage structure is symbolically correct, independent of the concrete geometric shape.
\subsection{\label{sec:geometry}Geometric Correctness}
The main utility of geometric correctness constraints is to compute the exact positions $\NN_i=\TWO{x_i}{y_i}$ of each node in the 2D workspace. These positions are functions of time $t$ and we sample a set of $T$ discrete time instances $t^{1,\cdots,T}$. In this section, we will always use superscripts for timestep indices. For example, at time instance $t^d$, the position of $\NN_i$ is $\NN_i^d$. We want to find a common geometric specification such that all the end-effector positions $\NN_K^{1,\cdots,T}$ can be achieved. 

The most important geometric variable is the length of each rigid rod. We define these parameters implicitly using a set of constraints such that, if $\NN_i$ and $\NN_j$ are connected, then the distance between these two nodes is a constant for all time instances. In other words, we need the following set of constraints if $C_{ji}=1$:
\begin{align}
\label{eq:equidistant_nonconvex}
\|\NN_j^d-\NN_i^d\|^2=\|\NN_j^{(d\bmod T)+1}-\NN_i^{(d\bmod T)+1}\|^2\quad\forall 1\leq d\leq T,
\end{align}
after which any distance $\|\NN_j^d-\NN_i^d\|^2$ can be used as the rigid rod length.

\begin{figure}[t]
\vspace{-10px}
  \begin{minipage}{.4\textwidth}
  \begin{equation}
  \begin{aligned}
  \label{eq:upper}
  \TWOC{\alpha}{\tilde{\alpha}}=\sum_{s=1}^S\lambda_s\TWOC{\alpha_s}{\alpha_s^2}& \\
  \{\lambda_{1,\cdots,S}\}\in\SOSB&  \\
  \sum_{s=1}^S\lambda_s=1&,
  \end{aligned}
  \end{equation}
  \end{minipage}\hfill\hspace{0.01\textwidth}
  \begin{minipage}{.5\textwidth}
  \includegraphics[width=0.99\textwidth]{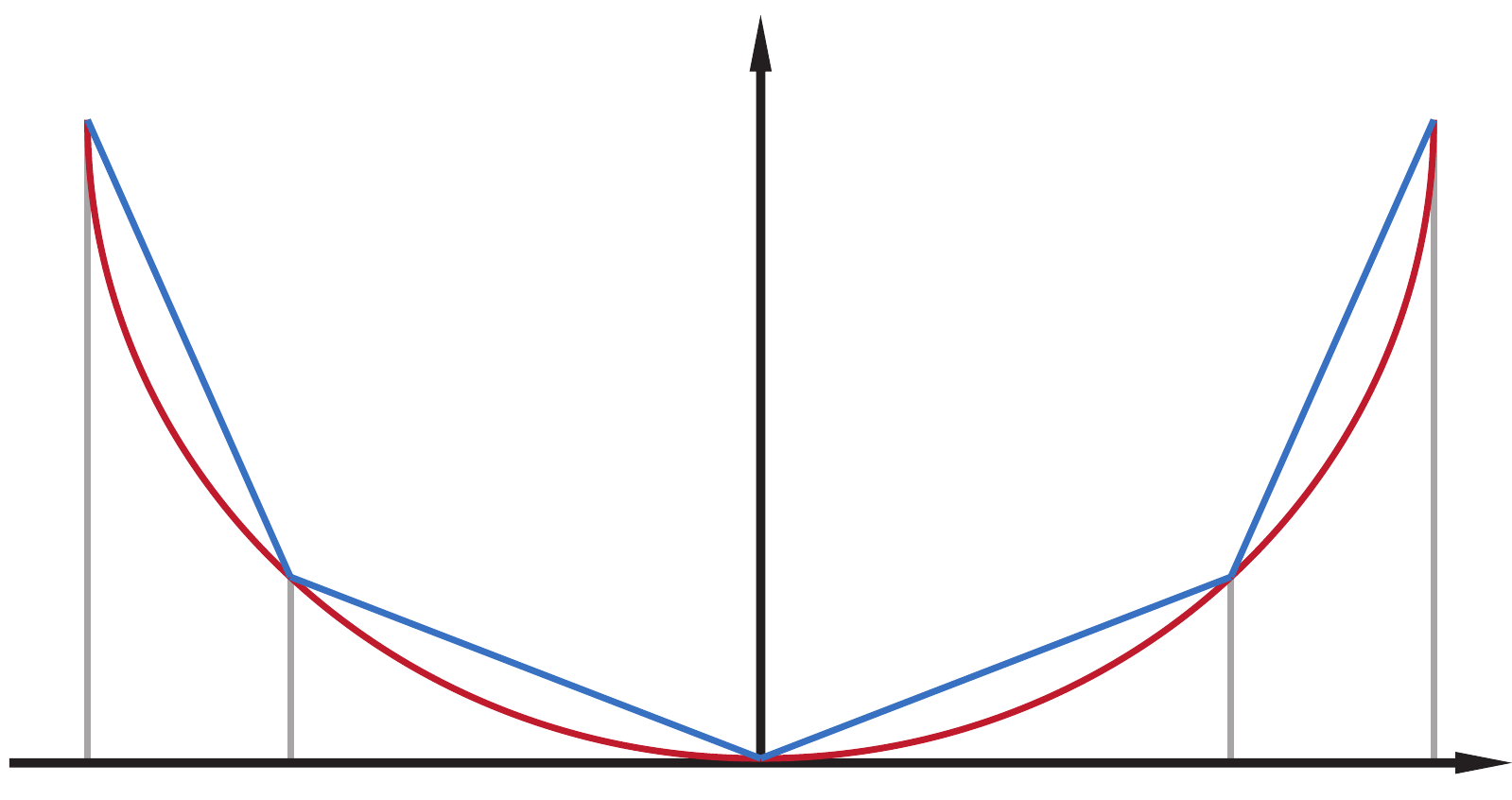}
  \put(-165,-5){$\alpha_1$}
  \put(-140,-5){$\alpha_2$}
  \put(-90 ,-5){$\alpha_3$}
  \put(-33 ,-5){$\alpha_4$}
  \put(-10 ,-5){$\alpha_5$}
  \put(-82 ,82){$\alpha^2$}
  \end{minipage}
\vspace{-5px}
\caption{\label{fig:upperBound} An illustration of the piecewise linear upper bound (blue) of the quadratic curve $\alpha^2$ (red) with $S=5$.}
\vspace{-5px}
\end{figure}
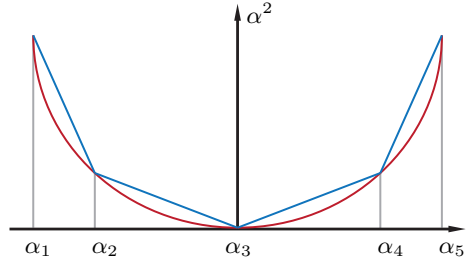

However, there are two challenging issues in modeling these constraints that can affect the performance of the MICP solver. A first challenge is to minimize the use of binary variables. Because any pair of nodes $\NN_j$ and $\NN_i$ might be connected, a naive formulation will require a number of binary variables proportional to $K^2$. Instead, we introduce auxiliary term $\E{d}_{1i}^d=\TWO{dx_{1i}^d}{dy_{1i}^d}$, which indicates the relative position between $\NN_i$ and the first other node connected to it at time instance $t^d$. Similarly, $\E{d}_{2i}^d=\TWO{dx_{2i}^d}{dy_{2i}^d}$ indicates the relative position between $\NN_i$ and the second other node connected to it. These definitions induce the following big-M constraints:
\begin{equation}
\begin{aligned}
\label{eq:distanceDef}
&\exists \{dx,dy\}_{ki}^d\quad\forall k=1,2\land i=2,\cdots,K\land d=1,\cdots,T    \\
&|\{dx,dy\}_{ki}^d-\{x,y\}_j^d+\{x,y\}_i^d|\leq 2B(1-C_{ji}^k)\quad\forall j=1,\cdots,i-1,
\end{aligned}
\end{equation}
where $B$ is the big-M parameter, implying that all the node positions lie in a bounded region $[-B,B]^2$. Note that the first motor node $\NN_1$ follows a circular curve (\prettyref{eq:motor}), which requires special definitions of $\E{d}_{11}^d,\E{d}_{21}^d$ as follows:
\begin{align}
\label{eq:distanceDefMotor}
\{dx,dy\}_{11}^d=\{dx,dy\}_{21}^d=\{x_1^d-X_C,y_1^d-Y_C\},
\end{align}
where the center of rotation $\TWO{X_C}{Y_C}$ is used as an additional auxiliary variable. The second challenge is that these constraints are non-convex because they involve quadratic terms. Fortunately, efficient formulations have been developed to relax non-convex functions using piecewise linear approximation \cite{liberti2004reformulation} and a special ordered set of type 2 ($\SOSB$) \cite{vielma2011modeling}. $\SOSB$ effects a constraint that at most two of the variables in an ordered set with consecutive indices can take non-zero values. To use these formulations, we decompose the range $[-B,B]$ evenly into $S-1$ pieces with $S$ nodes:
\begin{align*}
\{\alpha_i|-B=\alpha_1<\alpha_2<\cdots<\alpha_S=B\}.
\end{align*}
As a result, for any $\alpha\in[-B,B]$, a piecewise linear upper bound of $\alpha^2$ is $\tilde{\alpha}$, which is defined in \prettyref{eq:upper}.
As illustrated in \prettyref{fig:upperBound}, $\alpha^2\leq\tilde{\alpha}$ and this upper bound can be arbitrarily tight as $S\to\infty$. This formulation has been used in \cite{dai2017global} to discretize the space of unit vectors. In the rest of the paper, we use a tilde to denote such an upper bound. Using these upper bounds, the equidistant constraints can be approximated using the following conic constraints:
\begin{equation}
\begin{aligned}
\label{eq:equidistant}
&\forall i=1,\cdots,K\land d=1,\cdots,T \\
&\|\NN_i^d-\NN_i^{(d\bmod T)+1}\|^2\leq(2\sqrt{2}B)^2(1-F_i) \\
&\forall k=1,2\land i=1,\cdots,K\land d=1,\cdots,T \\
&\|\E{d}_{ki}^{(d\bmod T)+1}\|^2\leq \tilde{dx}_{ki}^d+\tilde{dy}_{ki}^d+(2\sqrt{2}B)^2F_i    \\
&\|\E{d}_{ki}^d\|^2\leq \tilde{dx}_{ki}^{(d\bmod T)+1}+\tilde{dy}_{ki}^{(d\bmod T)+1}+(2\sqrt{2}B)^2F_i
\end{aligned}
\end{equation}
where the last term on the right-hand sides is the big-M term that excludes fixed nodes. The idea is to require the length of two vectors to be smaller than the upper bound of one another. Note that \prettyref{eq:equidistant} converges to \prettyref{eq:equidistant_nonconvex} as $S\to\infty$. This formulation will require an upper bound for all $\E{d}_{ki}^d$ and each upper bound requires $\lceil\E{log}S\rceil$ binary variables. As a result, our formulation will introduce $\BO(4TK\lceil\E{log}S\rceil)$ binary decision variables. We also introduce a last constraint to ensure that rigid rods are not degenerate by ensuring minimal rod length $l_{min}$:
\begin{equation}
\begin{aligned}
\label{eq:miniLength}
&\forall k=1,2\land i=1,\cdots,K\land d=1,\cdots,T  \\
&\tilde{dx}_{ki}^d+\tilde{dy}_{ki}^d\geq l_{min}^2-((2\sqrt{2}B)^2+l_{min}^2)F_i.
\end{aligned}
\end{equation}

\begin{figure}[t]
\vspace{-5px}
\centering
\includegraphics[width=0.99\textwidth]{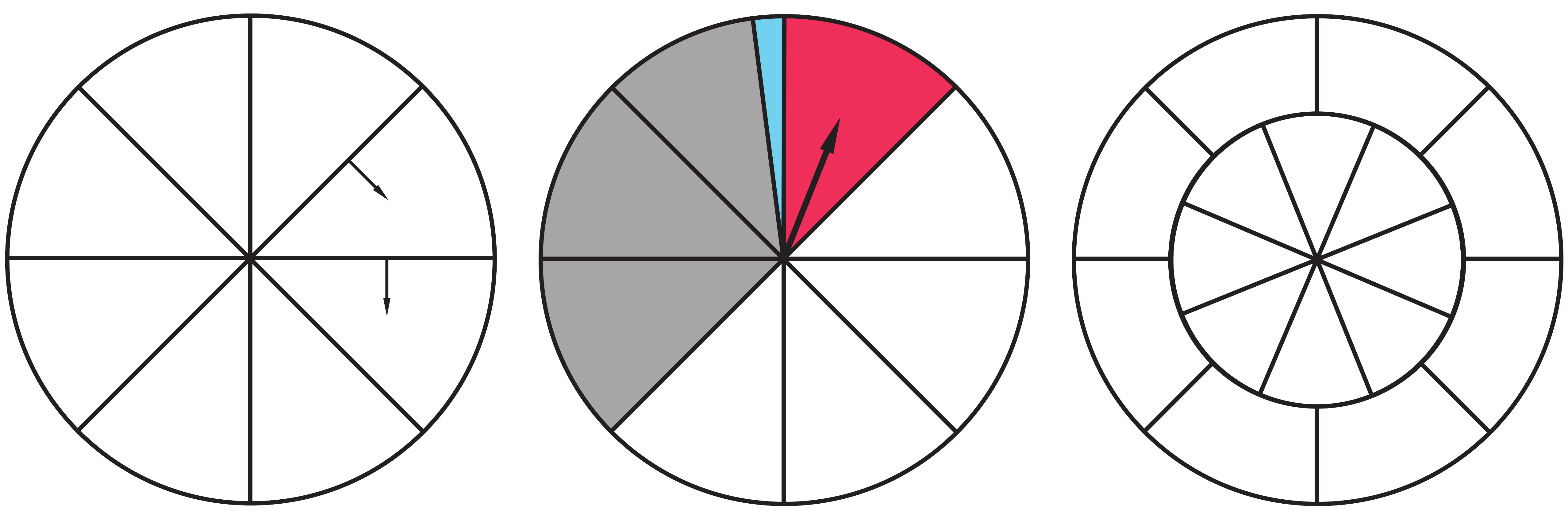}
\put(-270,60){$\E{v}_1^L$}
\put(-273,43){$\E{v}_1^R$}
\GAMMA{0}{292}{55}{45}{$\gamma_1$}
\GAMMA{1}{292}{55}{45}{$\gamma_2$}
\GAMMA{2}{292}{55}{45}{$\gamma_3$}
\GAMMA{3}{292}{55}{45}{$\gamma_4$}
\GAMMA{4}{292}{55}{45}{$\gamma_5$}
\GAMMA{5}{292}{55}{45}{$\gamma_6$}
\GAMMA{6}{292}{55}{45}{$\gamma_7$}
\GAMMA{7}{292}{55}{45}{$\gamma_8$}
\put(-295,-10){(a)}
\put(-178,113){$\epsilon$}
\put(-160, 90){$\E{d}_{1i}^d$}
\put(-210, 65){$\E{d}_{2i}^d$}
\put(-177,-10){(b)}
\GAMMA{0}{60}{55}{45}{$\gamma_1$}
\GAMMA{1}{60}{55}{45}{$\gamma_2$}
\GAMMA{2}{60}{55}{45}{$\gamma_3$}
\GAMMA{3}{60}{55}{45}{$\gamma_4$}
\GAMMA{4}{60}{55}{45}{$\gamma_5$}
\GAMMA{5}{60}{55}{45}{$\gamma_6$}
\GAMMA{6}{60}{55}{45}{$\gamma_7$}
\GAMMA{7}{60}{55}{45}{$\gamma_8$}
\GAMMAI{0}{60}{55}{45}{$\gamma_9$}
\GAMMAI{1}{60}{55}{45}{$\gamma_{10}$}
\GAMMAI{2}{60}{55}{45}{$\gamma_{11}$}
\GAMMAI{3}{60}{55}{45}{$\gamma_{12}$}
\GAMMAI{4}{60}{55}{45}{$\gamma_{13}$}
\GAMMAI{5}{60}{55}{45}{$\gamma_{14}$}
\GAMMAI{6}{60}{55}{45}{$\gamma_{15}$}
\GAMMAI{7}{60}{55}{45}{$\gamma_{16}$}
\put(-60,-10){(c)}
\vspace{-5px}
\caption{\label{fig:angleConstraint} Linear relaxation of angle constraints. (a): We cut $\mathcal{SO}(2)$ into 8 sectors, each of which is selected by a $\gamma$-flag using $\SOSA$ constraints. A sector, e.g. the sector selected by $\gamma_1$, is bounded by its left/right unit-length plane-normal vectors $\E{v}_1^L$/$\E{v}_1^R$. (b): If $\E{d}_{1i}^d$ falls in the red area, then we restrict $\E{d}_{2i}^d$ to its left half-space (gray), which is at least $\epsilon$-apart (blue). However, note that when $\E{d}_{1i}^d$ moves across sector boundaries, the gray area will jump discontinuously. (c): To avoid discontinuous changes in the restricted region for $\E{d}_{2i}^d$ when $\E{d}_{1i}^d$ undergoes continuous changes, we propose to double cover $\mathcal{SO}(2)$ using $2S=16$ sectors.}
\vspace{-15px}
\end{figure}
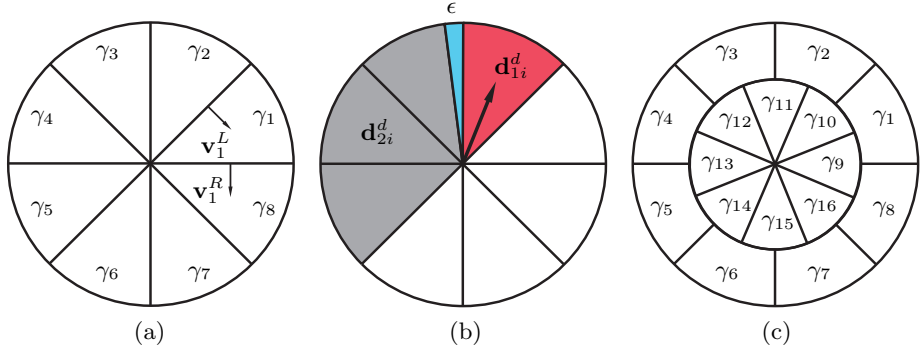
By ensuring a fixed rigid rod length across all time instances, we can make sure that all the end-effector positions $\NN_K^d$ can be achieved using the same planar linkage structure. In practice, however, we can only change the end-effector position by moving the first motor node $\NN_1$, so we still need to ensure that the mechanics system will not glitch or does not have singular configurations. The most intuitive classification of singular configuration is the rank-deficiency of the Jacobian matrix \cite{BOHIGAS20131}. However, this classification cannot be used in an MICP formulation because it is non-convex and the Jacobian matrix cannot be computed under our implicit representation of rigid rods. Instead, we adopt a heuristic proposed by \cite{Thomaszewski:2014:CDL:2601097.2601143}, which avoids singularities by ensuring that, for any movable node $\NN_i$, the two vectors $\E{d}_{1i}^d$ and $\E{d}_{2i}^d$ are not colinear. In other words, the triangle area formed by these two vectors is positive. This constraint takes the following bilinear form:
\begin{align*}
\E{d}_{1i}^d\times\E{d}_{2i}^d\geq\epsilon,
\end{align*}
where $\epsilon$ is a small constant. Although this constraint is bilinear, we can use McCormick envelopes \cite{liberti2004reformulation} to relax it as a conic constraint. If the range $[-B,B]$ is cut into $S-1$ segments, then this formulation will introduce $\BO(4TK\lceil\E{log}S\rceil)$. However, a critical flaw of this formulation is that a McCormick envelope is an outer-approximation. As a result, the exact linkage structure can still be singular, although its conic relaxation is non-singular. To ensure strict non-singular formulation, we propose a constraint whereby the angle between the two vectors is larger than $\epsilon$, which is equivalent to the positive area constraint when combined with \prettyref{eq:miniLength}:
\begin{equation}
\begin{aligned}
\label{eq:singular_nonconvex}
\measuredangle\E{d}_{1i}^d,\E{d}_{2i}^d\geq\epsilon.
\end{aligned}
\end{equation}
In addition, we propose an inner approximation such that the exact linkage structure is also guaranteed to be non-singular. Concretely, we cut the space of $\mathcal{SO}(2)$ into $S$ sectors, as illustrated in \prettyref{fig:angleConstraint}a, so that $\E{d}_{1i}^d$ will only fall into one of the $S$ sectors. If $\E{d}_{1i}^d$ falls in a particular sector, then we restrict $\E{d}_{2i}^d$ to its left half-space that is at least $\epsilon$-apart, as shown in \prettyref{fig:angleConstraint}b. If we use an $\SOSA$ constraint to select the sector in which $\E{d}_{1i}^d$ falls, then only $\BO(TK\lceil\E{log}S\rceil)$ binary decision variables are needed. A minor issue with this formulation is that the allowed region of $\E{d}_{2i}^d$ jumps discontinuously as $\E{d}_{1i}^d$ changes continuously. We can fix this problem by double-covering the region of $\mathcal{SO}(2)$ using $2S$ sectors, as shown in \prettyref{fig:angleConstraint}c, which will introduce $\BO(TK\lceil\E{log}2S\rceil)$ binary decision variables.

To formulate these constraints, we assume that each sector of $\mathcal{SO}(2)$ is flagged by a selector variable $\gamma_l$, which is bounded by its left/right unit-length plane-normal vectors $\E{v}_l^L$/$\E{v}_l^R$. Then the following constraints must be satisfied if $\E{d}_{1i}^d$ falls inside the sector:
\begin{equation}
\begin{aligned}
\label{eq:sector}
<\E{v}_l^L,\E{d}_{1i}^d>\geq 0&\\
<\E{v}_l^R,\E{d}_{1i}^d>\leq 0& \\
<\E{R}(\epsilon)\E{v}_l^L,\E{d}_{2i}^d>\leq 0&  \\
<\E{R}(\pi)\E{v}_l^R,\E{d}_{2i}^d>\geq 0&,
\end{aligned}
\end{equation}
where $\E{R}(\bullet)$ is the $2\times2$ counter-clockwise rotation matrix by angle $\bullet$. Combined with the fact that \prettyref{eq:sector} should only be satisfied for one particular sector and that only movable nodes satisfy these constraints, we have the following formulation:
\begin{equation}
\begin{aligned}
\label{eq:sectorFinal}
\forall i=2,\cdots,K\quad d=1,\cdots,T&    \\
<\E{v}_l^L,\E{d}_{1i}^d>\geq 2\sqrt{2}B(\gamma_{l,i}^d-1)-2\sqrt{2}BF_i& \\
<\E{v}_l^R,\E{d}_{1i}^d>\leq 2\sqrt{2}B(1-\gamma_{l,i}^d)+2\sqrt{2}BF_i& \\
<\E{R}(\epsilon)\E{v}_l^L,\E{d}_{2i}^d>\leq 2\sqrt{2}B(1-\gamma_{l,i}^d)+2\sqrt{2}BF_i&  \\
<\E{R}(\pi)\E{v}_l^R,\E{d}_{2i}^d>\geq 2\sqrt{2}B(\gamma_{l,i}^d-1)-2\sqrt{2}BF_i&  \\
\{(\gamma_{1,1i}^d,\cdots,\gamma_{2S,i}^d\}\in\SOSA& \\
\sum_{l=1}^{2S}\gamma_{l,i}^d=1&.
\end{aligned}
\end{equation}
These constraints will avoid singular configurations.
\subsection{The Complete MICP Formulation}
Combining all the constraints, we minimize two objective function terms. First, we want the end-effector trajectory to match the target trajectory specified by users. Second, to minimize manufacturing cost, we want to use as few rigid rods as possible. To formulate the first objective term, we need to replace a trajectory with a discrete number of samples. However, the order of these samples is discarded. In practice, we find that better solutions can be found by preserving the order between these samples. This requirement is formulated by making sure that $\NN_K^d$ will be visited by the end-effector sequentially when the motor node rotates by $2\pi$ either clockwise or counter-clockwise. This requirement is formulated using the following MICP constraints:
\begin{equation}
\begin{aligned}
\label{eq:rotation}
&\forall d=1,\cdots,T-1 \\
\|\E{R}(\frac{2\pi}{T})\E{d}_{11}^d-\E{d}_{11}^{d+1}\|^2&\leq(2\sqrt{2}B)^2D&   \\
\|\E{R}(-\frac{2\pi}{T})\E{d}_{11}^d-\E{d}_{11}^{d+1}\|^2&\leq(2\sqrt{2}B)^2(1-D),
\end{aligned}
\end{equation}
where $D$ is a binary variable to choose which direction the motor rotates. Putting everything together, we arrive at the following MICP problem:
\begin{equation}
\begin{aligned}
\label{eq:MICP}
\argmin{}&\sum_{d=1}^T\|\NN_K^d-\NN_K^{d*}\|^2+w\sum_{i=1}^K U_i    \\
\argcon{}&\text{\prettyref{eq:state}, \ref{eq:connectivity}, \ref{eq:balance}, \ref{eq:balance2},
\ref{eq:distanceDef}, \ref{eq:distanceDefMotor}, \ref{eq:equidistant},
\ref{eq:miniLength}, \ref{eq:sectorFinal}, \ref{eq:rotation},}
\end{aligned}
\end{equation}
where $\NN_K^{d*}$ are the sampled points on the target trajectory and $w$ the regularization weight of the cost-efficiency term. Since non-convexity is not accepted by MICP, the solution returned by MICP is only a piecewise linear approximation of the original nonlinear problem. To return a solution with exact constraint satisfaction, we refine the solution by solving an additional NLP locally using the following formulation:
\begin{equation}
\begin{aligned}
\label{eq:NLP}
\argmin{}&\sum_{d=1}^T\|\NN_K^d-\NN_K^{d*}\|^2    \\
\argcon{}&\text{\prettyref{eq:state}, \ref{eq:connectivity}, \ref{eq:balance}, \ref{eq:balance2},
\ref{eq:distanceDef}, \ref{eq:distanceDefMotor}, \ref{eq:equidistant_nonconvex}, 
\ref{eq:singular_nonconvex}, \ref{eq:rotation},}
\end{aligned}
\end{equation}
where we fix all the binary variables $U_i,F_i,D$. Note that \prettyref{eq:NLP} is a mixed-integer NLP (MINLP) generalization of \prettyref{eq:MICP} and we have the following lemma:
\begin{lemma}
\prettyref{eq:MICP} converges to \prettyref{eq:NLP} as $S\to\infty$, and the BB algorithm can find the global optimum for \prettyref{eq:MICP}.
\end{lemma}
\section{\label{sec:results}Results and Evaluations}
\begin{figure}[t]
\centering
\includegraphics[width=0.99\textwidth]{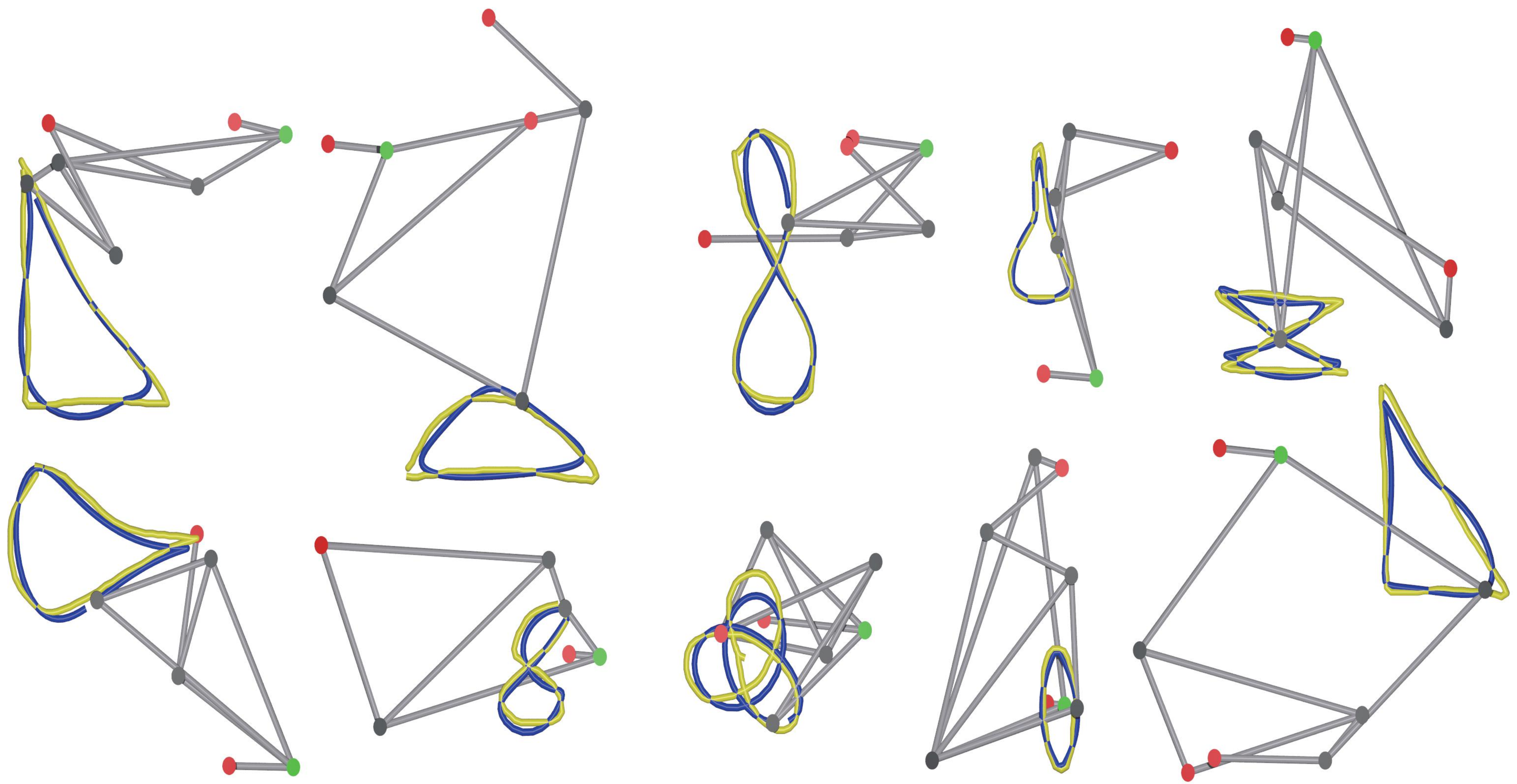}
\caption{\label{fig:results_gallary} We show 10 different optimized planar linkage structures with the end-effector trajectory in blue and the user-specified target trajectory in yellow. For all these examples, we choose $K=5\sim7$, $S=9$, and $T=10\sim20$. The end-effector trajectory matches closely with the target trajectory.}
\vspace{-10px}
\end{figure}

We have implemented our method using Gurobi \cite{gurobi} as our MICP solver for \prettyref{eq:MICP} and Knitro \cite{byrd2006k} as our NLP solver for \prettyref{eq:NLP}. All the experiments are performed on a cluster with 4 CPU cores per process (2.5GHz E5-2680 CPU). Compared with prior work \cite{Thomaszewski:2014:CDL:2601097.2601143}, the main benefit of our formulation is that we can search for planar linkage structures from a target trajectory of the end-effector that requires trivial effort from users. In \prettyref{fig:results_gallary}, we show a list of different target trajectories and the optimized planar linkage structures. 

\begin{figure}[h]
\centering
\includegraphics[width=0.32\textwidth]{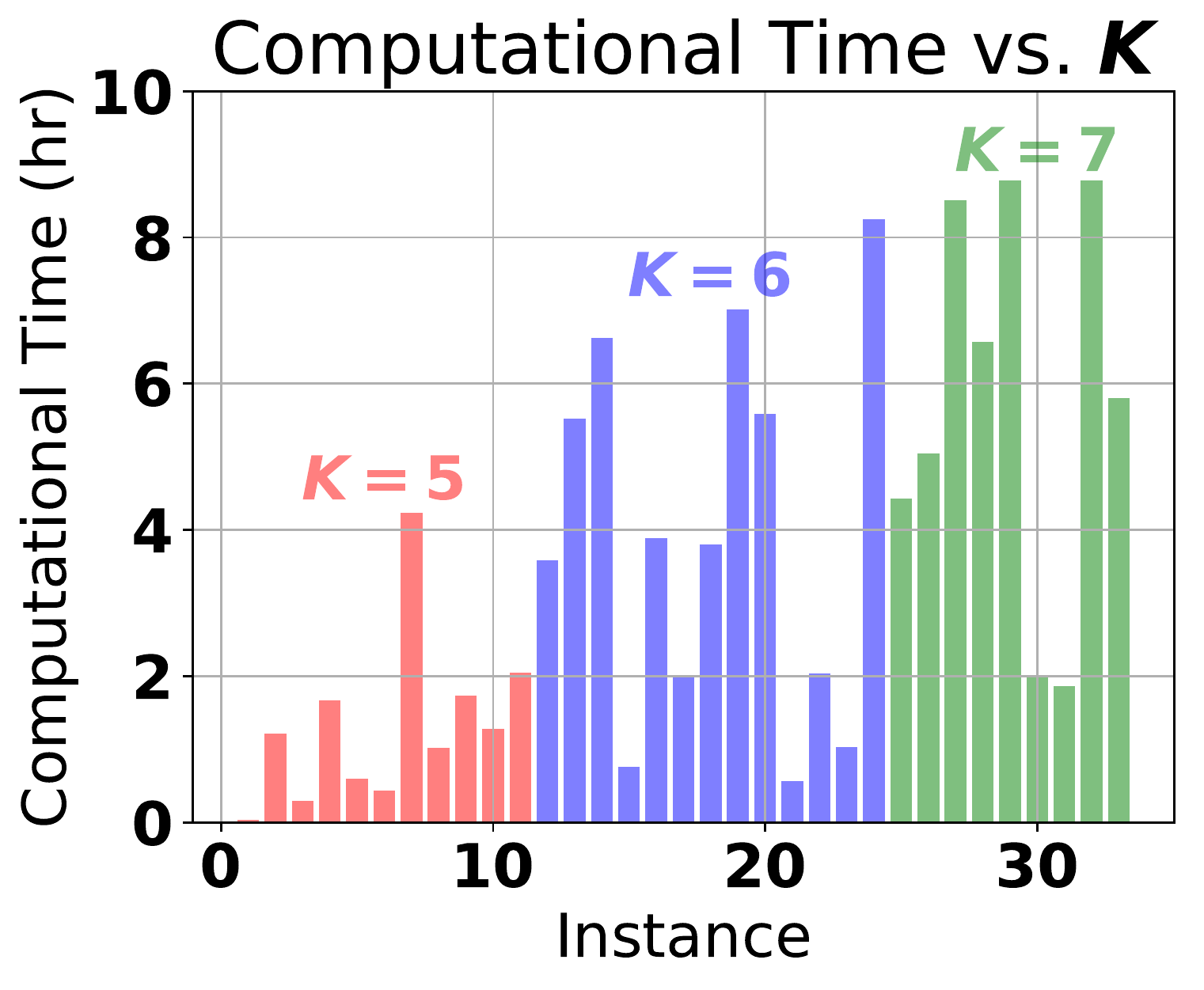}
\includegraphics[width=0.32\textwidth]{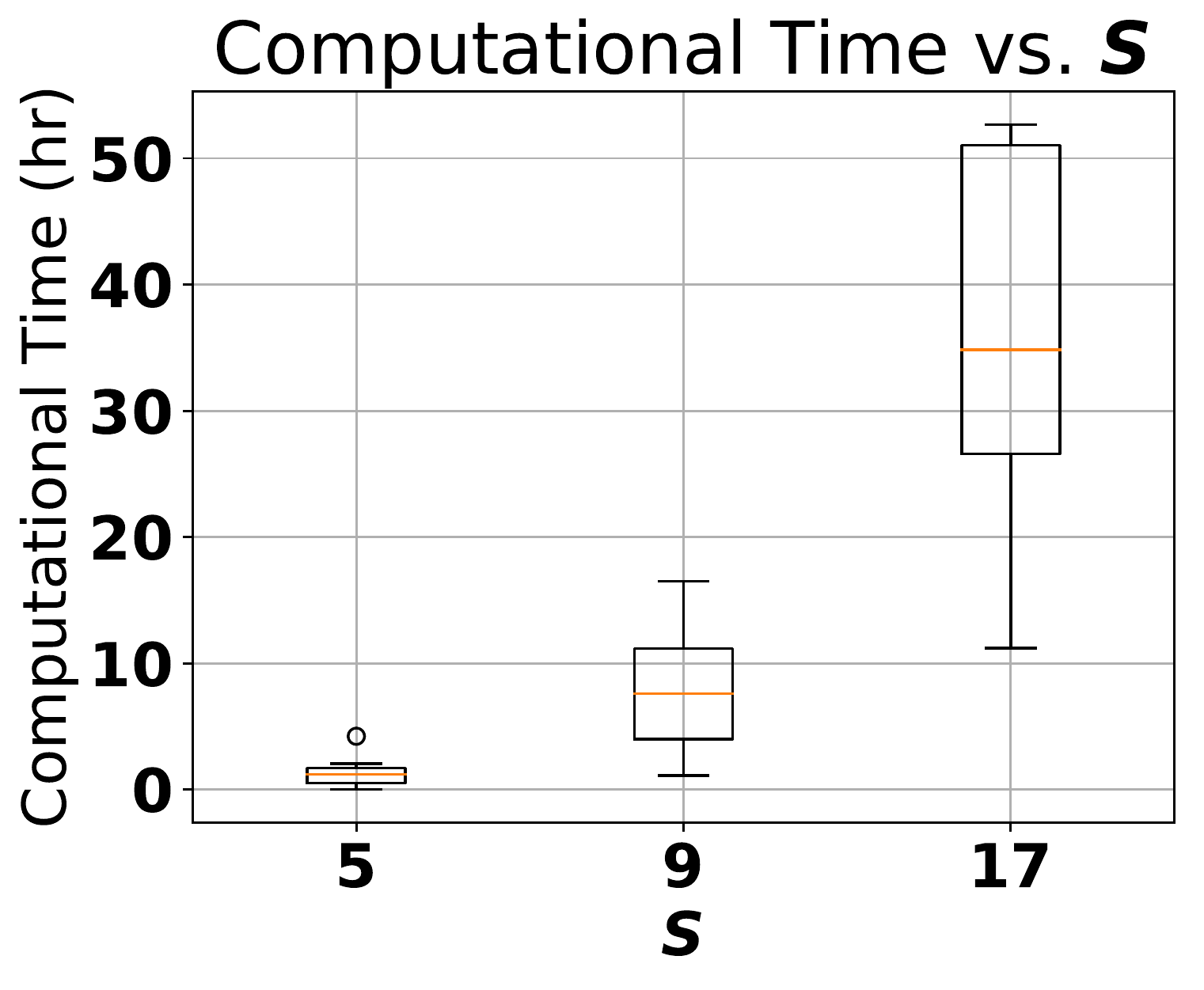}
\includegraphics[width=0.32\textwidth]{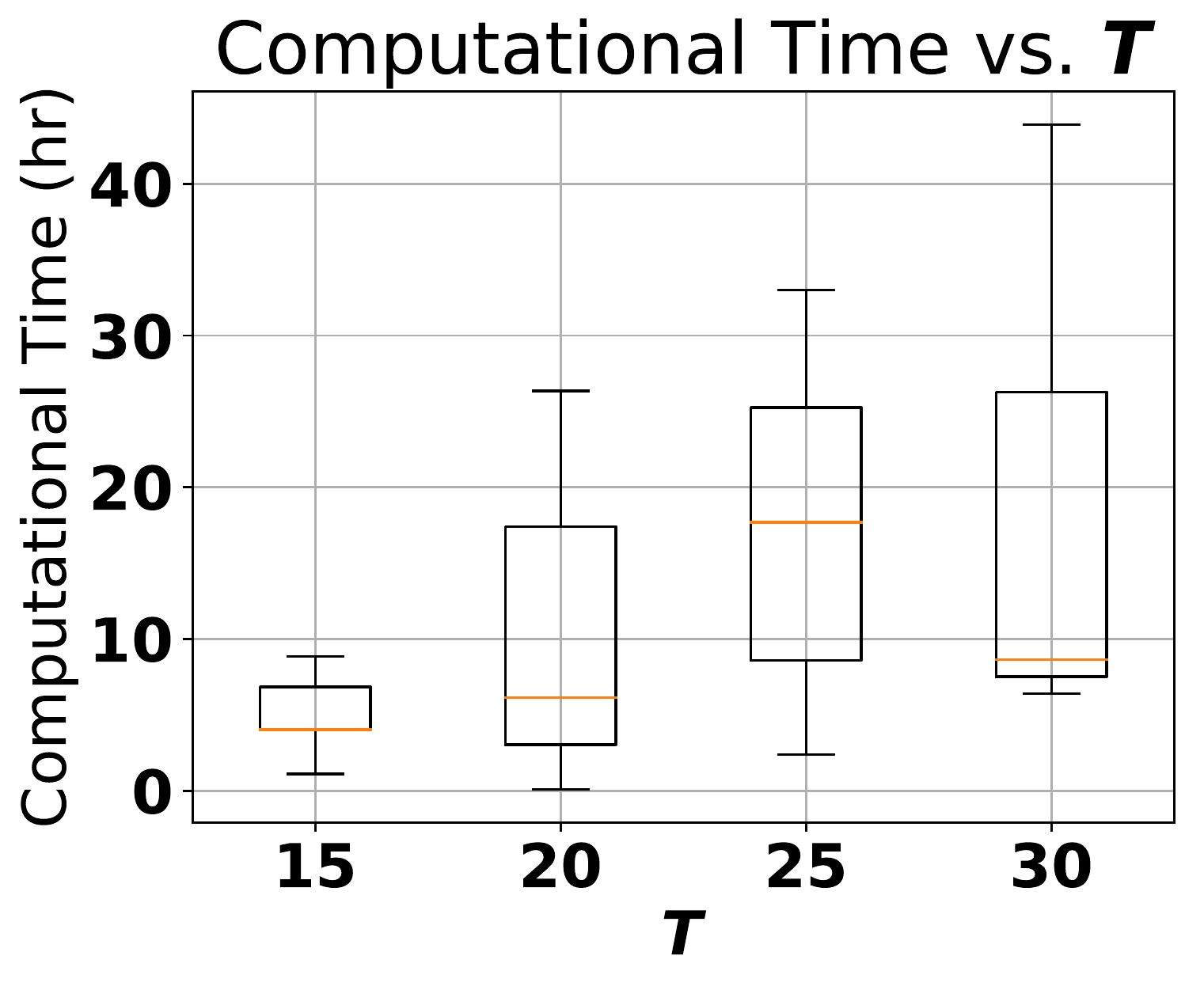}
\vspace{-5px}
\caption{\label{fig:scalability} We plot the average computational time for solving MICP in 35 example problems using different parameters in \prettyref{fig:results_gallary}. The computational time for solving MICP grows exponentially with $K$, $\E{log}S$, and $T$.}
\vspace{-15px}
\end{figure}
\begin{figure}[h]
\vspace{-10px}
\centering
\includegraphics[width=0.31\textwidth]{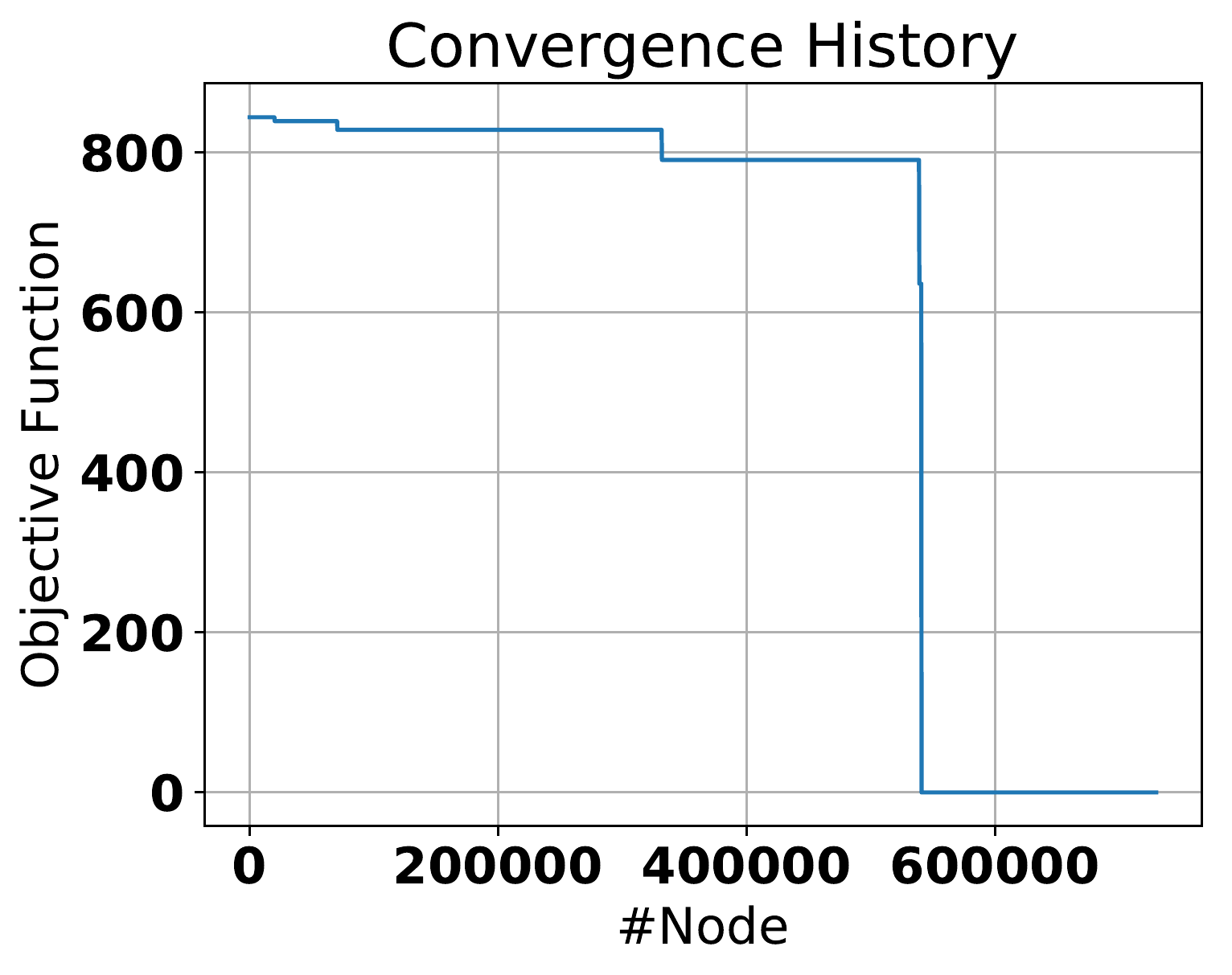}
\includegraphics[width=0.31\textwidth]{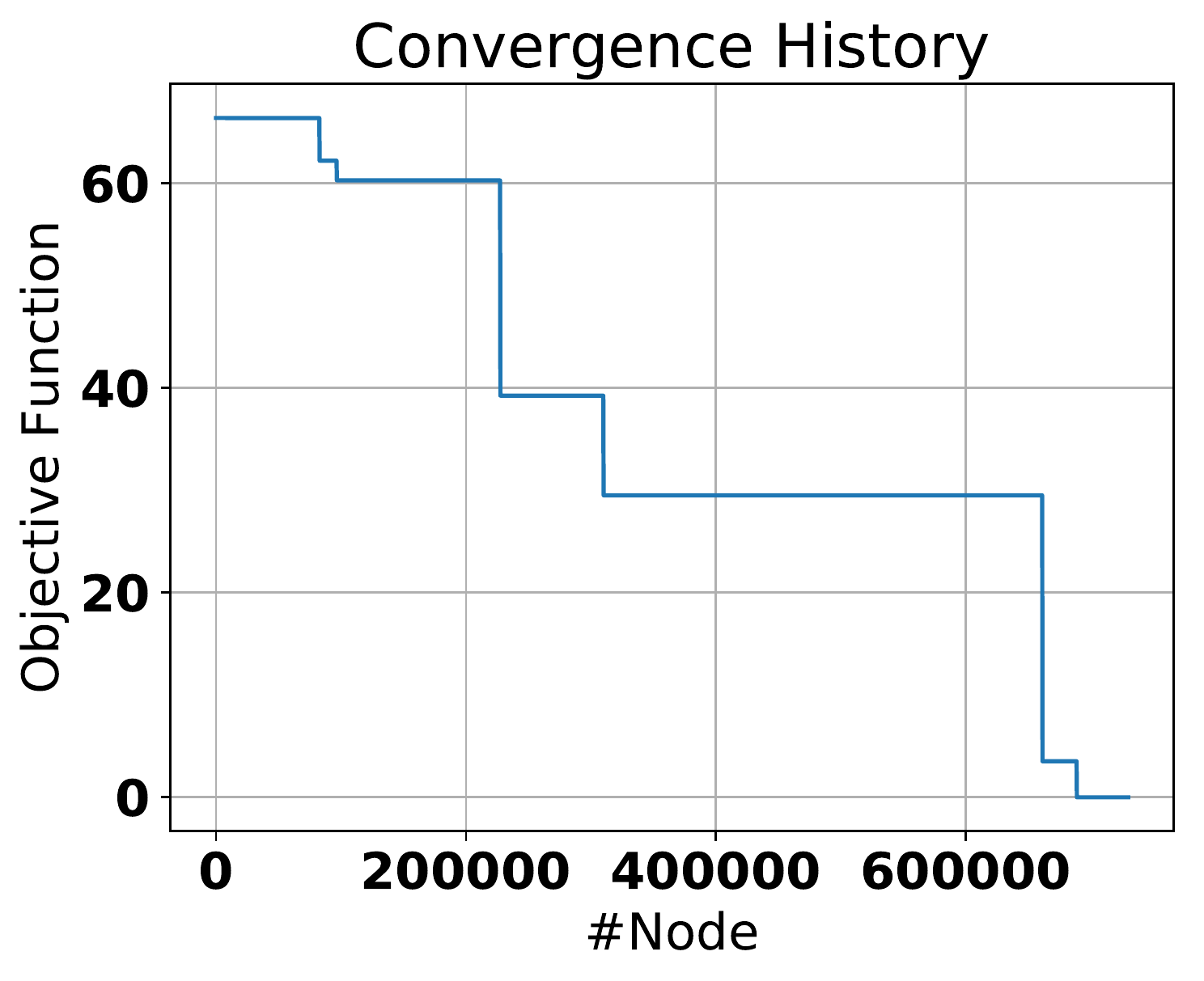}
\includegraphics[width=0.31\textwidth]{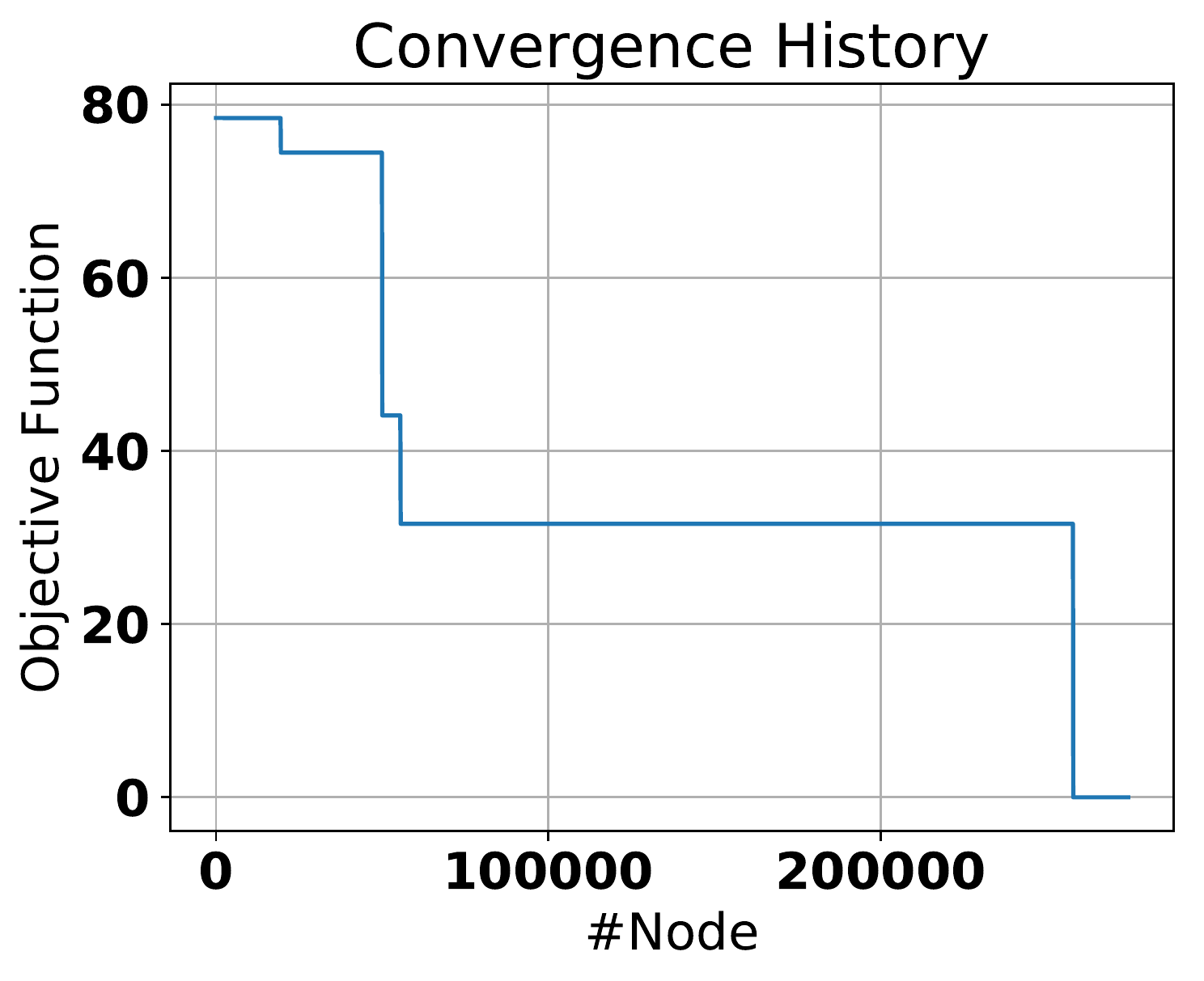}
\vspace{-5px}
\caption{\label{fig:optimal_gap} We plot the convergence history curve for 3 typical optimizations by showing the objective function values plotted against the number of nodes explored in the BB search tree. The BB algorithm spends most of its time exploring infeasible nodes and the first identified feasible solution is usually very close to the optimal solution, so that the optimizer will return the globally optimal solution after refining the solution for $5-10$ times.}
\vspace{-15px}
\end{figure}
The performance and accuracy of our algorithm heavily depend on the three parameters: the max number of rigid rods $K$, the number of pieces for approximating $S$, and the number of samples on the target trajectory $T$. Since the cost of solving MICP grows exponentially with the number of binary decision variables, which is proportional to $K$, $\E{log}S$, and $T$, our method cannot scale to large problems, as illustrated in \prettyref{fig:scalability}. In practice, we find that, given a maximal computational time of $10$ hours, we can compute globally optimal solutions for most benchmarks with $K\leq 7$, $S\leq 9$, and $T\leq 20$. This is enough if we design robots part-by-part, as is done in the Theo Jansen's strandbeest. For other benchmarks, the computational time is longer than $10$ hours, but a feasible solution has been found, although it is sub-optimal. In \prettyref{fig:optimal_gap}, we plot the average convergence history of a typical optimization. Since we express all the topology and geometric requirements as hard mixed-integer constraints, feasible solutions are quite rare in the search space and the optimizer takes most of the computational time pruning infeasible solutions. Once the first feasible solution is found, it is usually very close to the optimal solution and the optimizer refines it for less than 10 times to reach the optimal solution. 

\begin{figure}[h]
\vspace{-10px}
\centering
\includegraphics[width=0.9\textwidth]{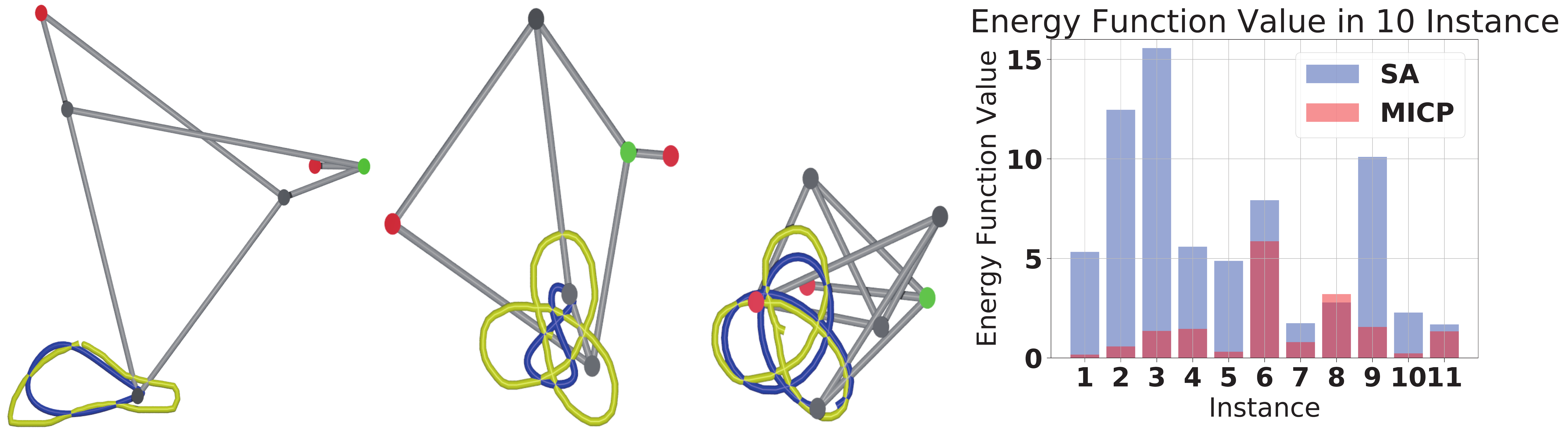}
\put(-270,0){(a)}
\put(-220,0){(b)}
\put(-140,0){(c)}
\put(-20 ,0){(d)}
\vspace{-5px}
\caption{\label{fig:SAComparison} SA can find good enough solutions for simple target curves (a). However, for more complex curve shapes, SA failes (b) while MICP succeeds (c). We also plot the objective function values returned by SA and MICP in 10 computational examples in (d), where MICP outperforms SA in 9 instances.}
\vspace{-15px}
\end{figure}
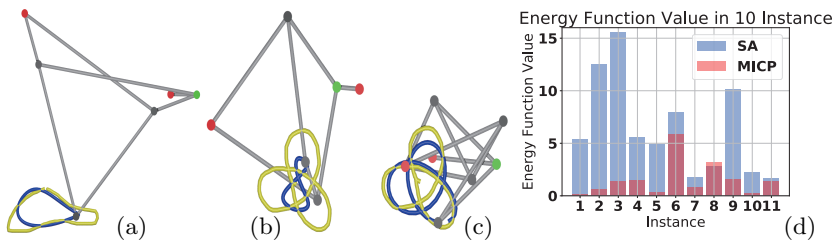
We have also compared our method with conventional global search algorithms such as simulated annealing (SA). We implemented a similar algorithm as proposed in \cite{Zhu:2012:MMT:2366145.2366146}. In this algorithm, we randomly generate $1000000$ samples by random moves and accept these samples according to the simulated annealing rule. Each random move can be of one of three kinds: geometric change, node addition, and node removal. In geometric change, the length of a rigid rod is randomly perturbed. In node addition, a new node is added and the length of the new rigid rods are randomly picked. In node removal, the end-effector node is removed and the last movable node is used as the new end-effector node. We enhance standard SA algorithm by making sure that each random move is valid. In other words, we introduce an inner loop and repeated try random moves until the modified planar linkage structure satisfies all the topological constraints and has no singular configurations. As illustrated in \prettyref{fig:SAComparison}a, SA algorithm can find satisfactory results for simple target curves, but SA usually fails for more complex curve shapes (\prettyref{fig:SAComparison}bc). In \prettyref{fig:SAComparison}d, we also show the objective function values after convergence. The solution of MICP is almost always better than the solution of SA. However, SA outperforms MICP in one example, which is probably due to the inexact constraint satisfaction of MICP. 

\begin{figure}[h]
\vspace{-10px}
\centering
\scalebox{0.9}{
\includegraphics[width=0.9\textwidth]{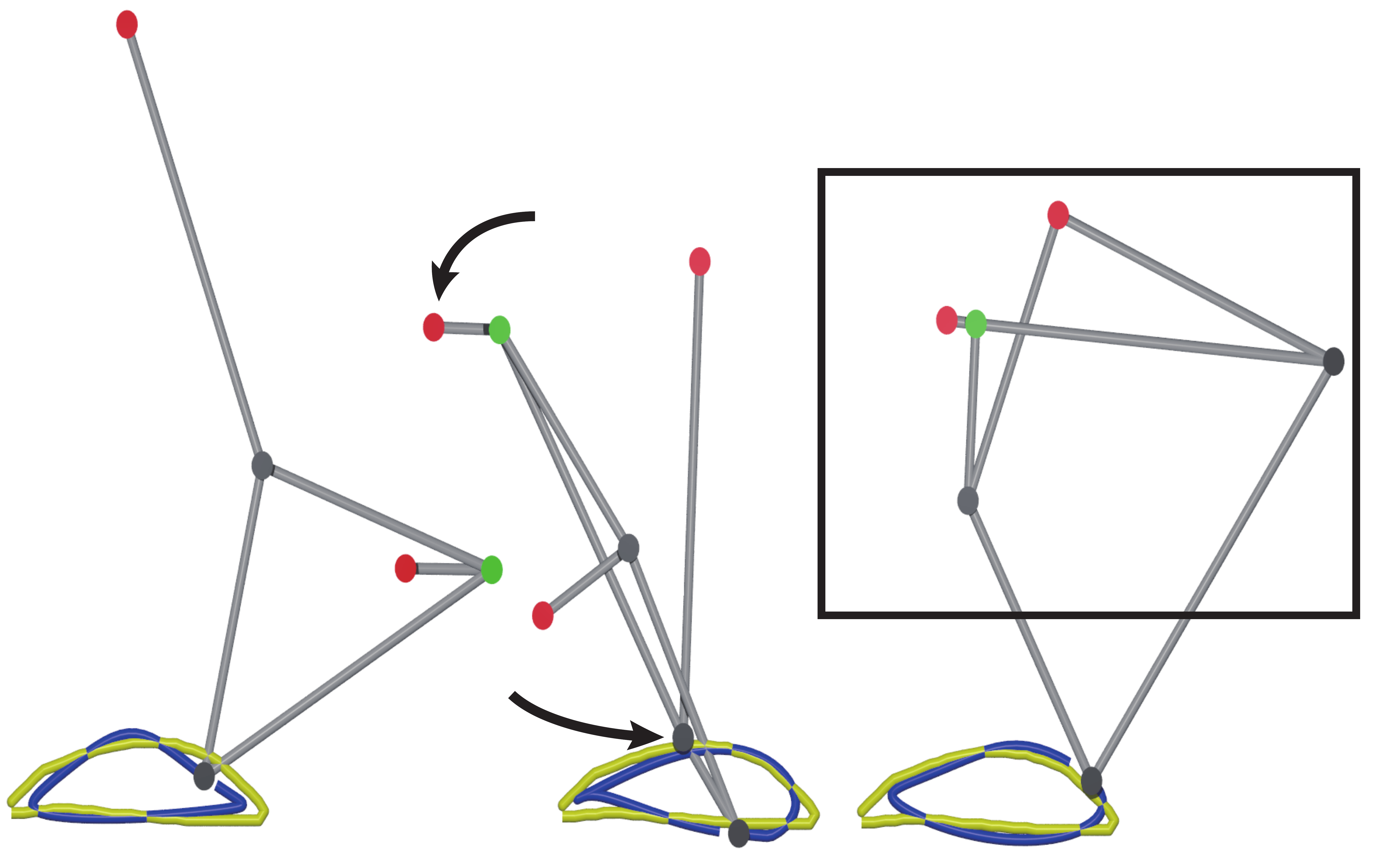}
\put(-240,10){(a)}
\put(-210,10){(b)}
\put(-50 ,10){(c)}
\put(-185,145){Fixed}
\put(-220,45){Too Close}
\put(-100,165){Bounded Region}}
\vspace{-10px}
\caption{\label{fig:constraints} We show results taking two different user constraints into consideration. (a): Results with no constraints. The optimizer is guided by the regularization term to use as few nodes as possible. (b): We fix the center of rotation and the optimizer finds a more complex structure with 6 nodes. (c): If we do not want the structure to be too close to the target curve, we can add a bounded region and create a constraint that any nodes (other than the end-effector node) should be inside the bounded region.}
\vspace{-15px}
\end{figure}
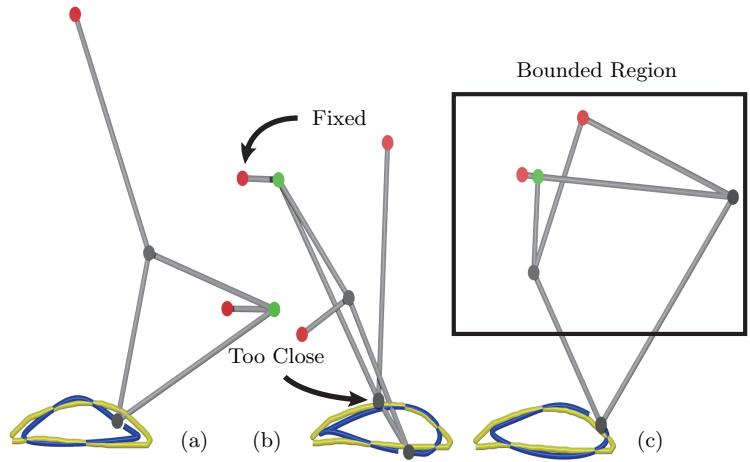

Usually, the design of a planar linkage structure is not only subject to a target end-effector trajectory, but also to various other user constraints. For example, the user might require certain nodes to be fixed, which can be easily achieved using our MICP formulation. The user may also reserve certain parts of the robot for some functional units that cannot be occupied by the planar linkages. This type of constraint can be expressed as collision avoidance between a planar linkage structure and a specified convex region, which can be formulated as MICP constraints using a prior method \cite{ding2011mixed}. In \prettyref{fig:constraints}, we show results taking these constraints into consideration.
\section{Conclusion \& Limitations}
We present a globally optimal formulation to jointly search for both the topology and geometry of a planar linkage structure. Our formulation relaxes the problem into a MICP, for which optimal solutions can be found efficiently using BB algorithms. Our results show that our formulation can search for complex structures from trivial and intuitive user inputs, i.e. target end-effector trajectories. Additionally, various design constraints can be easily incorporated. For moderately complex structures, the solve time using these formulations falls in the range between minutes and hours. 

\begin{figure}[t]
\centering
\vspace{-5px}
\includegraphics[width=0.9\textwidth]{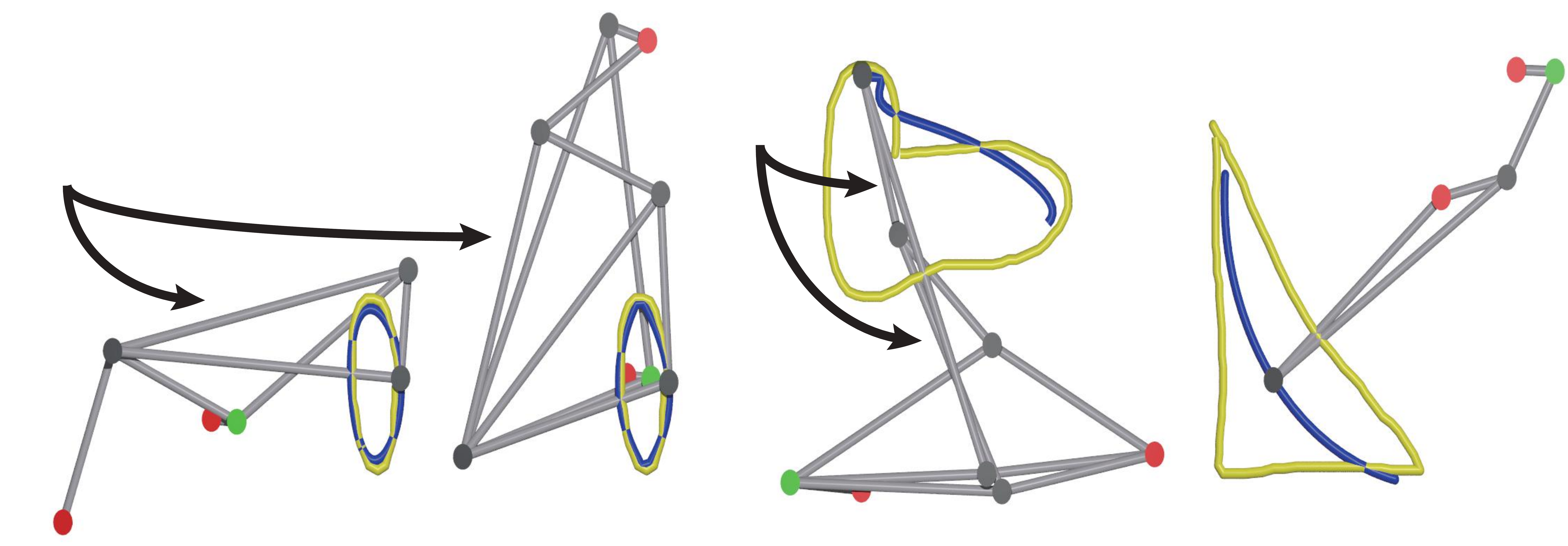}
\put(-280,0){(a)}
\put(-200,0){(b)}
\put(-120,0){(c)}
\put(-50 ,0){(d)}
\put(-320,75){Ambiguity}
\put(-185,82){Colinear}
\put(-80,90){Sub-Optimal}
\caption{\label{fig:failcase} Failure cases and issues with our formulation. (a,b): MICP only returns the single global optimum. But similar target trajectories can lead to two different linkage structures. (c): We only satisfy geometric constraints approximately, so that the linkage structure might not satisfy these constraints exactly. In this example, we have two rigid rods being colinear. (d): Usually, the early feasible solutions found by MICP are of low-quality, and we have to wait for the MICP to find the global optimum.}
\end{figure}
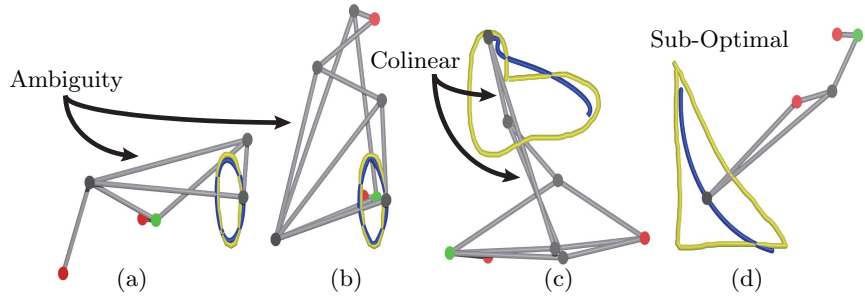

As a major limitation, the solve time increases quickly with the number of possible rigid bodies in the planar structure ($K$) and the number of samples on the target trajectory ($T$) because the number of decision variables depends on a multiplication of these two parameters. A related issue is that MICP only satisfies the geometric constraints approximately. As illustrated in \prettyref{fig:failcase}c, a predicted target trajectory with approximate constraints satisfaction can be different from a predicted target trajectory with exact constraints satisfaction after solving \prettyref{eq:NLP}. To reduce the approximation error, we have to increase the approximation granularity by using a larger $S$, which in turn increases the number of binary decision variables. Finally, note that our formulation does not generate all possible planar linkage structures but only those that allow sequential forward kinematic processing. This problem is inherited from \cite{kecskemethy1997symbolic,bacher2015linkedit} by using the same representation as these works. Allowing more general planar linkages is also possible under the MICP formulation by using a new formulation of topology constraints.

\subsection{Future Work}
Our future research will focus on a balance between global optimality and formulation efficiency. Such a balance could possibly be achieved by using MINLP formulations. In addition, we observe that different planar linkages, as shown in \prettyref{fig:failcase}ab, can generate very similar target trajectories. This indicates that there exist many local optima with objective function close to the global optimum. However, a BB algorithm will only return the single global optimum. In addition, we found that we need to wait until the BB algorithm finds its global optimum; the intermediary solutions are not usually usable, as illustrated in \prettyref{fig:failcase}d. A potential future direction is to use algorithms such as Bayesian optimization that can explore multiple local optima and return many solutions for users to make a choice.

\bibliographystyle{splncs03}
\bibliography{ref}

\end{document}